\documentclass[11pt,letterpaper]{article}
\usepackage{emnlp2017}
\usepackage{times}
\usepackage{latexsym}
\usepackage{url}

\usepackage{capt-of}
\usepackage{tabularx}
\usepackage{amsmath}
\usepackage{multirow}
\usepackage[caption=false]{subfig}
\usepackage{epsfig}
\usepackage{amssymb}
\usepackage{amsmath}
\usepackage{amsfonts}
\usepackage[english]{babel}
\usepackage{blindtext}
\usepackage{booktabs}

\newcommand{\tabincell}[2]{\begin{tabular}{@{}#1@{}}#2\end{tabular}}
\newcommand{\tx}[1]{``\textit{#1}''}

\DeclareMathOperator\sigmoid{\textrm{sigmoid}}

\DeclareMathOperator\avg{\textrm{avg}}

\DeclareMathOperator\maximize{\textrm{maximize}}

\DeclareMathOperator*{\argmax}{arg\,max}

\DeclareMathOperator*{\lstm}{\textrm{LSTM}}
\definecolor{myred}{rgb}{0.8, 0.0, 0.0}

\emnlpfinalcopy

\title{Learning to Paraphrase for Question Answering}

\author{Li Dong$^{\dagger}$ \and Jonathan Mallinson$^{\dagger}$ \and Siva Reddy$^{\ddagger}$ \and Mirella Lapata$^{\dagger}$ \\
	$^{\dagger}$ ILCC, School of Informatics, University of Edinburgh \\
	$^{\ddagger}$ Computer Science Department, Stanford University \\
	{\tt li.dong@ed.ac.uk},~~{\tt J.Mallinson@ed.ac.uk} \\
	{\tt sivar@stanford.edu},~~{\tt mlap@inf.ed.ac.uk}}

\date{}

\begin{document}

\maketitle

\begin{abstract}
Question answering (QA) systems are sensitive to the many different ways natural language expresses the same information need. In this paper we turn to paraphrases as a means of capturing this knowledge and present a general framework which learns felicitous paraphrases for various QA tasks. Our method is trained end-to-end using question-answer pairs as a supervision signal. A question and its paraphrases serve as input to a neural scoring model which assigns higher weights to linguistic expressions most likely to yield correct answers. We evaluate our approach on QA over Freebase and answer sentence selection. Experimental results on three datasets show that our framework consistently improves performance, achieving competitive results despite the use of simple QA models.
\end{abstract}

\section{Introduction}

Enabling computers to automatically answer questions posed in natural
language on any domain or topic has been the focus of much research in
recent years. Question answering (QA) is challenging due to the many
different ways natural language expresses the same information
need. As a result, small variations in semantically equivalent
questions, may yield different answers. For example, a hypothetical QA
system must recognize that the questions \tx{who created microsoft}
and \tx{who started microsoft} have the same meaning and that they
both convey the \texttt{founder} relation in order to retrieve the
correct answer from a knowledge base.

\begin{figure*}[t]
  \centering
  \includegraphics[width=1\textwidth]{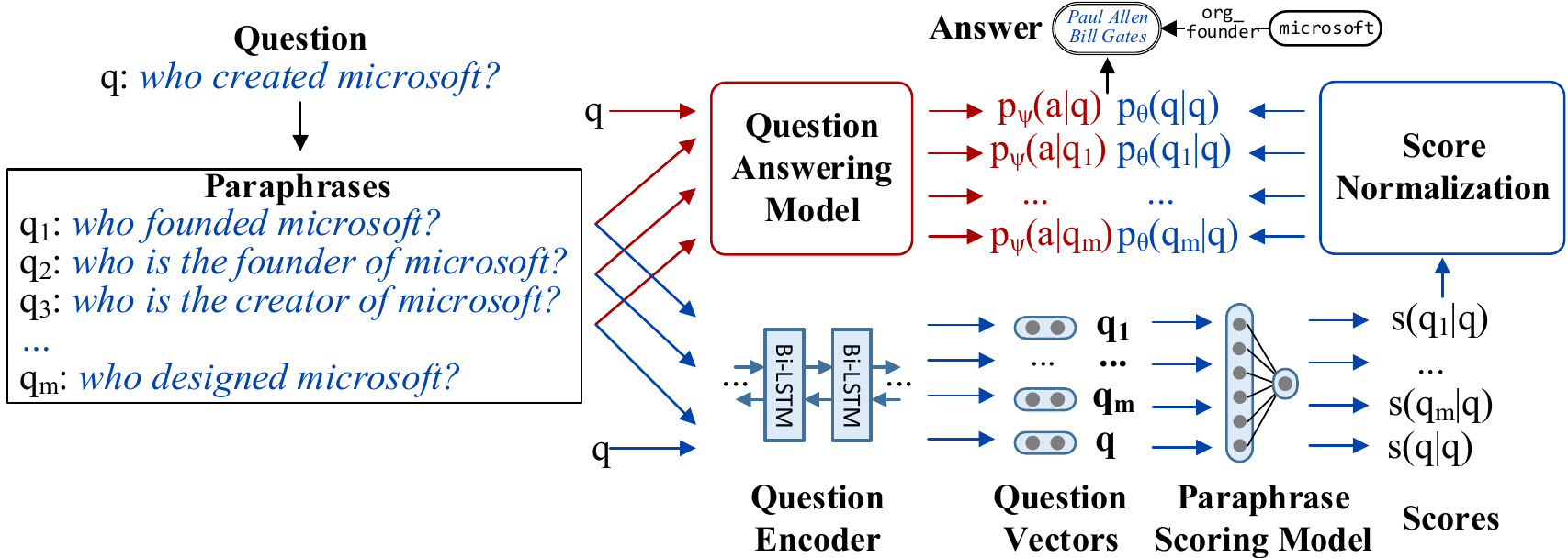}
  \caption{We use three different methods to generate candidate
    paraphrases for input~$q$. The question and its paraphrases are
    fed into a neural model which scores how suitable they are. The
    scores are normalized and used to weight the results of the
    question answering model. The entire system is trained end-to-end
    using question-answer pairs as a supervision signal.}
  \label{fig:method}
\end{figure*}

Given the great variety of surface forms for semantically equivalent
expressions, it should come as no surprise that previous work has
investigated the use of paraphrases in relation to question
answering. There have been three main strands of research. The first
one applies paraphrasing to match natural language and logical forms in the context of semantic parsing.
\citet{parasempre} use a template-based method to heuristically
generate canonical text descriptions for candidate logical forms, and
then compute paraphrase scores between the generated texts and input
questions in order to rank the logical forms.
Another strand of work uses paraphrases in the context of neural
question answering models~\cite{qa:subgraph,qa:embedding:model,qa:mccnn}. These models are typically trained on
question-answer pairs, and employ question paraphrases in a multi-task
learning framework in an attempt to encourage the neural networks to
output similar vector representations for the paraphrases.

The third strand of research uses paraphrases more directly. The idea
is to paraphrase the question and then submit the rewritten version to
a QA module.  Various resources have been used to produce question
paraphrases, such as rule-based machine
translation~\cite{answer:question:you:wished}, lexical and phrasal
rules from the Paraphrase Database \cite{latent-pcfg-sp}, as well as
rules mined from Wiktionary \cite{rewrite-sp} and large-scale
paraphrase corpora~\cite{openqa13}.  A common problem with the
generated paraphrases is that they often contain inappropriate
candidates.  Hence, treating all paraphrases as equally felicitous and
using them to answer the question could degrade performance. To remedy
this, a scoring model is often employed, however independently of the
QA system used to find the
answer~\cite{answer:question:you:wished,latent-pcfg-sp}.
Problematically, the separate paraphrase models used in previous work
do not fully utilize the supervision signal from the training data,
and as such cannot be properly tuned to the question answering tasks
at hand. Based on the large variety of possible transformations that
can generate paraphrases, it seems likely that the kinds of
paraphrases that are useful would depend on the QA application of
interest~\cite{what:is:paraphrase}.  \citet{openqa14} use features
that are defined over the original question and its rewrites to score
paraphrases. Examples include the pointwise mutual information of the
rewrite rule, the paraphrase's score according to a language model,
and POS tag features.  In the context of semantic parsing,
\citet{rewrite-sp} also use the ID of the rewrite rule as a feature.
However, most of these features are not informative enough to model
the quality of question paraphrases, or cannot easily generalize to
unseen rewrite rules.

In this paper, we present a general framework for learning paraphrases
for question answering tasks. Given a natural language question, our
model estimates a probability distribution over candidate answers. We
first generate paraphrases for the question, which can be obtained by
one or several paraphrasing systems. A neural scoring model predicts
the quality of the generated paraphrases, while learning to assign
higher weights to those which are more likely to yield correct
answers. The paraphrases and the original question are fed into a QA
model that predicts a distribution over answers given the question.
The
entire system is trained end-to-end using question-answer pairs as a
supervision signal. The framework is flexible, it does not rely on
specific paraphrase or QA models. In fact, this plug-and-play
functionality allows to learn specific paraphrases for different QA
tasks and to explore the merits of different paraphrasing models
for different applications. 

We evaluate our approach on QA over Freebase and
text-based answer sentence selection. We employ a range of paraphrase
models based on the Paraphrase Database (PPDB; \citealt{ppdb2}),
neural machine translation \cite{jnmt}, and rules mined from the
WikiAnswers corpus \cite{openqa14}.  Results on three datasets show
that our framework consistently improves performance; it achieves
state-of-the-art results on GraphQuestions and competitive performance
on two additional benchmark datasets using simple QA models.

\section{Problem Formulation}

Let~$q$ denote a natural language question, and~$a$ its answer. Our
aim is to estimate~$p\left(a | q\right)$, the conditional probability
of candidate answers given the question.  We decompose~$p\left(a |
  q\right)$ as:
\begin{equation}
\label{eq:overview}
\hspace*{-.2cm}p\left(a | q\right) = \sum_{q' \in {H_q \cup \{q\} }} { \underbrace{p_{\psi}\left(a | q' \right)}_{\text{~~~~QA Model~~~~}} \underbrace{p_{\theta}\left(q' | q\right)}_{\text{Paraphrase Model}} }
\end{equation}
where $H_q$ is the set of paraphrases for question~$q$, $\psi$ are the
parameters of a QA model, and $\theta$~are the parameters of a paraphrase
scoring model.

As shown in Figure~\ref{fig:method}, we first generate candidate
paraphrases $H_q$ for question~$q$.  Then, a neural scoring model
predicts the quality of the generated paraphrases, and assigns higher
weights to the paraphrases which are more likely to obtain the correct
answers.  These paraphrases and the original question simultaneously
serve as input to a QA model that predicts a distribution over answers
for a given question.  Finally, the results of these two models are
fused to predict the answer.
In the following we will explain how~$p\left(q' | q\right)$ and
$p\left(a | q'\right)$~are estimated.

\subsection{Paraphrase Generation}
\label{sec:paraphr-gener}

As shown in Equation~\eqref{eq:overview}, the term $p\left(a |
  q\right)$ is the sum over $q$ and its paraphrases $H_q$. Ideally, we
would generate all the paraphrases of~$q$. However, since this set
could quickly become intractable, we restrict the number of candidate
paraphrases to a manageable size.  In order to increase the coverage
and diversity of paraphrases, we employ three methods based on:
(1)~lexical and phrasal rules from the Paraphrase Database
\cite{ppdb2}; (2)~neural machine translation
models~\cite{seq2seq,mt:jointly:align:translate}; and (3)~paraphrase
rules mined from clusters of related questions \cite{openqa14}.  We
briefly describe these models below, however, there is nothing
inherent in our framework that is specific to these, any other
paraphrase generator could be used instead.

\begin{table}[t]
	\fontsize{8}{8.5}\selectfont
	\centering
	\setlength\tabcolsep{2pt}
	\begin{tabular}{@{}l@{~}r@{}}
		\toprule
		\multicolumn{2}{l}{\textbf{Input}: {what be the zip code of the largest car manufacturer}} \\ \hline
		{what be the zip code of the largest vehicle manufacturer}       & PPDB                    \\
		{what be the zip code of the largest car producer}               & PPDB                    \\
		{what be the postal code of the biggest automobile manufacturer} & NMT                     \\
		{what be the postcode of the biggest car manufacturer}           & NMT                     \\
		{what be the largest car manufacturer 's postal code}            & Rule                    \\
		{zip code of the largest car manufacturer}                       & Rule                    \\ \bottomrule
	\end{tabular}
	\caption{Paraphrases obtained for an input question from different
		models (PPDB, NMT, Rule). Words are lowercased and stemmed.} 
	\label{table:para:example}
\end{table}

\subsubsection{PPDB-based Generation}
\label{sec:para:ppdb}

Bilingual pivoting \cite{bilingual:pivot} is one of the most
well-known approaches to paraphrasing; it uses bilingual parallel
corpora to learn paraphrases based on techniques from phrase-based
statistical machine translation (SMT, \citealt{koehn2003statistical}).
The intuition is that two English strings that translate to the same
foreign string can be assumed to have the same meaning. The method
first extracts a bilingual phrase table and then obtains English
paraphrases by pivoting through foreign language phrases.

Drawing inspiration from syntax-based SMT, \citet{ppdb:2008} and
\citet{ppdb:2011} extended this idea to syntactic paraphrases,
leading to the creation of PPDB~\cite{ppdb1}, a large-scale
paraphrase database containing over a billion of paraphrase pairs
in $24$ different languages. \citet{ppdb2} further used a supervised
model to automatically label paraphrase pairs with entailment
relationships based on natural logic \cite{MacCartney:2009}.  In our
work, we employ bidirectionally entailing rules from
PPDB. Specifically, we focus on lexical (single word) and phrasal
(multiword) rules which we use to paraphrase questions by replacing
words and phrases in them. An example is shown in
Table~\ref{table:para:example} where we substitute \textit{car} with
\textit{vehicle} and \textit{manufacturer} with \textit{producer}.

\subsubsection{NMT-based Generation}
\label{sec:para:nmt}

\citet{jnmt} revisit bilingual pivoting in the context of neural
machine translation (NMT,
\citealt{seq2seq,mt:jointly:align:translate}) and present a
paraphrasing model based on neural networks. At its core, NMT
is trained end-to-end to maximize the
conditional probability of a correct translation given a source
sentence, using a bilingual corpus. Paraphrases can be obtained by
translating an English string into a foreign language and then
back-translating it into English. NMT-based pivoting models offer
advantages over conventional methods such as the ability to learn
continuous representations and to consider wider context while
paraphrasing.

\begin{figure}[t]
  \centering
\hspace*{-.1cm}\includegraphics[width=0.5\textwidth]{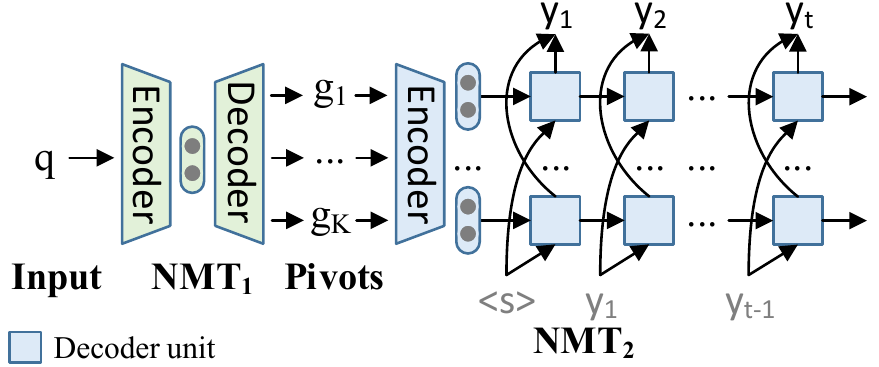}
\caption{Overview of NMT-based paraphrase generation. \textsc{NMT}$_1$
  (green) translates question~$q$ into pivots $g_1 \dots g_K$ which
  are then back-translated by \textsc{NMT}$_2$ (blue) where $K$
  decoders jointly predict tokens at each time step, rather than only
  conditioning on one pivot and independently predicting outputs.}
  \label{fig:jnmt}
\end{figure}

In our work, we select German as our pivot following \citet{jnmt}
who show that it outperforms other languages in a wide range of
paraphrasing experiments, and pretrain two NMT systems,
English-to-German (\textsc{En-De}) and German-to-English
(\textsc{De-En}).  A naive implementation would translate a question
to a German string and then back-translate it to English. However,
using only one pivot can lead to inaccuracies as it places too much
faith on a single translation which may be wrong. Instead, we
translate from multiple pivot sentences \cite{jnmt}. As shown in
Figure~\ref{fig:jnmt}, question~$q$ is translated to $K$-best German
pivots,~$\mathcal{G}_q = \{ g_1 , \dots, g_K \}$. The probability of
generating paraphrase \mbox{$q' = y_1 \dots y_{|q'|}$} is decomposed as:
\begin{equation}
\label{eq:jnmt}
\begin{aligned}
p\left( q' | \mathcal{G}_q \right) &= \prod _{ t=1 }^{ |q'| }{ p\left( y_t | y_{<t} , \mathcal{G}_q \right) } \\
&= \prod _{ t=1 }^{ |q'| }{ \sum _{ k=1 }^{ K }{ p\left( g_k | q \right) p\left( y_{ t }|y_{ <t }, g_k \right) } }
\end{aligned}
\end{equation}
where $y_{<t} = y_1 , \dots, y_{t-1}$, and $|q'|$ is the length
of~$q'$. Probabilities~$p\left( g_k | q \right)$ and~$p\left( y_{ t
  }|y_{ <t }, g_k \right)$ are computed by the \textsc{En-De} and
\textsc{De-En} models, respectively. We use beam search to decode
tokens by conditioning on multiple pivoting sentences. The results
with the best decoding scores are considered candidate paraphrases.
Examples of NMT paraphrases are shown in
Table~\ref{table:para:example}.

Compared to PPDB, NMT-based paraphrases are syntax-agnostic, operating
on the surface level without knowledge of any underlying
grammar. Furthermore, paraphrase rules are captured implicitly and
cannot be easily extracted, e.g., from a phrase table. As mentioned
earlier, the NMT-based approach has the potential of performing major
rewrites as paraphrases are generated while considering wider
contextual information, whereas PPDB paraphrases are more local, and
mainly handle lexical variation.

\subsubsection{Rule-Based Generation}
\label{sec:para:rule}

\begin{table}[t]
\fontsize{8.5}{9.5}\selectfont
  \centering
  \setlength\tabcolsep{1.5pt}
  \begin{tabular}{@{}l l@{}}
    \toprule
    \textbf{Source}                 & \textbf{Target}                     \\ \hline
    the average size of \_\_          & what be \_\_ average size             \\
    \_\_ be locate on which continent & what continent be \_\_ a part of      \\
    language speak in \_\_            & what be the official language of \_\_ \\
    what be the money in \_\_         & what currency do \_\_ use             \\ \bottomrule
  \end{tabular}
  \caption{Examples of rules used in the rule-based paraphrase generator.}
  \label{table:rule:example}
\end{table}

Our third paraphrase generation approach uses rules mined from the
WikiAnswers corpus~\cite{openqa14} which contains more than
$30$~million question clusters labeled as paraphrases by
WikiAnswers\footnote{\url{wiki.answers.com}} users.  This
corpus is a large resource (the average cluster size is~$25$), but is
relatively noisy due to its collaborative nature -- $45\%$ of question
pairs are merely related rather than genuine paraphrases.
We therefore followed the method proposed in \cite{openqa13} to
harvest paraphrase rules from the corpus.  We first extracted question
templates (i.e., questions with at most one wild-card) that appear in
at least ten clusters.  Any two templates co-occurring (more than five
times) in the same cluster and with the same arguments were deemed
paraphrases. Table~\ref{table:rule:example} shows examples of rules
extracted from the corpus.  During paraphrase generation, we consider
substrings of the input question as arguments, and match them with the
mined template pairs.  For example, the stemmed input question in
Table~\ref{table:para:example} can be paraphrased using the rules
(\tx{what be the zip code of \_\_}, \tx{what be \_\_ 's postal code}) and
(\tx{what be the zip code of \_\_}, \tx{zip code of \_\_}).  If no exact
match is found, we perform fuzzy matching by ignoring stop words in
the question and templates.

\subsection{Paraphrase Scoring}

Recall from Equation~\eqref{eq:overview} that $p_{\theta}\left(q' |
  q\right)$ scores the generated paraphrases $q' \in H_q \cup
\{q\}$. We estimate~$p_{\theta}\left(q' | q\right)$ using neural
networks given their successful application to paraphrase
identification
tasks~\cite{para:socher2011,para:cnn:yin,para:cnn:he}. Specifically,
the input question and its paraphrases are encoded as vectors. Then,
we employ a neural network to obtain the score~$s\left(q' | q\right)$
which after normalization becomes the probability~$p_{\theta}\left(q'
  | q\right)$.

\paragraph{Encoding} Let $q = q_1 \dots q_{|q|}$ denote an input
question.  Every word is initially mapped to a \mbox{$d$-dimensional}
vector. In other words, vector $\mathbf{q}_t$ is computed via
$\mathbf{q}_t = \mathbf{W}_q \mathbf{e}\left( q_t \right) $, where
$\mathbf{W}_q \in \mathbb{R}^{d \times |\mathcal{V}|}$ is a word
embedding matrix, $|\mathcal{V}|$~is the vocabulary size, and
$\mathbf{e}\left( q_t \right)$~is a one-hot
vector. 
Next, we use a bi-directional recurrent neural network with long
short-term memory units (\textsc{LSTM}, \citealt{lstm}) as the
question encoder, which is shared by the input questions and their
paraphrases. The encoder recursively processes tokens one by one, and
uses the encoded vectors to represent questions. We compute the hidden
vectors at the $t$-th time step via:
\begin{equation}
\begin{aligned}
\overrightarrow{\mathbf{h} }_{ t } &= \lstm\left( \overrightarrow{\mathbf{h} }_{ t-1 } , \mathbf{q}_t \right) , t = 1, \dots , |q| \\
\overleftarrow{\mathbf{h} }_{ t } &= \lstm\left( \overleftarrow{\mathbf{h} }_{ t+1 } , \mathbf{q}_t \right) , t = |q|, \dots , 1
\end{aligned}
\end{equation}
where $\overrightarrow{\mathbf{h} }_{ t }, \overleftarrow{\mathbf{h}
}_{ t } \in \mathbb{R}^{n}$. In this work we follow the $\lstm$
function described in~\citet{rnn:dropout}.  The representation
of~$q$ is obtained by:
\begin{equation}
\label{eq:question:vector}
\mathbf{q} = \left[ \overrightarrow{\mathbf{h} }_{ |q| } , \overleftarrow{\mathbf{h} }_{ 1 } \right]
\end{equation}
where $[\cdot,\cdot]$ denotes concatenation, and $\mathbf{q} \in
\mathbb{R}^{2n}$.

\paragraph{Scoring} After obtaining vector representations for~$q$ and
$q'$, we compute the score $s\left(q' | q\right)$ via:
\begin{equation}
\label{eq:score}
s\left(q' | q\right) = \mathbf{w}_s \cdot \left[ \mathbf{q} , \mathbf{q'} , \mathbf{q} \odot \mathbf{q'} \right] + b_s
\end{equation}
where $\mathbf{w}_s \in \mathbb{R}^{6n}$ is a parameter vector,
$[\cdot,\cdot,\cdot]$~denotes concatenation, $\odot$~is element-wise
multiplication, and $b_s$ is the bias.  Alternative ways to
compute~$s\left(q' | q\right)$ such as dot product or with a bilinear
term were not empirically better than Equation~\eqref{eq:score} and we
omit them from further discussion.

\paragraph{Normalization}
For paraphrases $q' \in H_q \cup \{q\}$, the probability
$p_{\theta}\left(q' | q\right)$ is computed via:
\begin{equation}
\label{eq:normalization}
p_{\theta}\left(q' | q\right) = \frac{\exp \{s\left(q' | q\right)\}}{\sum _{ r \in H_q \cup \{q\} }{ \exp \{s\left(r | q\right)\} }}
\end{equation}
where the paraphrase scores are normalized over the set $H_q \cup
\{q\}$.

\subsection{QA Models}
\label{sec:qa:model}

The framework defined in Equation~\eqref{eq:overview} is relatively
flexible with respect to the QA model being employed as long as it can
predict~$p_{\psi}\left(a | q' \right)$. We illustrate this by
performing experiments across different tasks and describe below the
models used for these tasks.

\paragraph{Knowledge Base QA}
In our first task we use the Freebase knowledge base to answer
questions. Query graphs for the questions
typically contain more than one predicate. For example, to answer the
question \tx{who is the ceo of microsoft in 2008}, we need to use one
relation to query \tx{ceo of microsoft} and another relation for the
constraint \tx{in 2008}.  For this task, we employ the
\textsc{SimpleGraph} model described
in~\citet{depgraphparser,reddy2017universal}, and follow their
training protocol and feature design.  In brief, their method uses
rules to convert questions to ungrounded
logical forms, which are subsequently matched against Freebase
subgraphs.  The QA model learns from question-answer pairs: it
extracts features for pairs of questions and Freebase subgraphs, and
uses a logistic regression classifier to predict the probability that
a candidate answer is correct.
We perform entity linking using the Freebasee/KG API on the original question~\cite{depgraphparser,reddy2017universal}, and generate candidate Freebase subgraphs.
The QA model estimates how likely it is for a subgraph to yield the correct answer.

\paragraph{Answer Sentence Selection}
Given a question and a collection of relevant sentences, the goal of
this task is to select sentences which contain an answer to the
question.  The assumption is that correct answer sentences have high
semantic similarity to the questions~\cite{bigramcnn,wikiqa,nasm}. We
use two bi-directional recurrent neural networks (\textsc{BiLSTM}) 
to separately encode questions and answer sentences to vectors
(Equation~\eqref{eq:question:vector}).
Similarity scores are computed as shown in Equation~\eqref{eq:score}, 
and then squashed to $(0,1)$ by a $\sigmoid$ function in order to predict $p_{\psi}\left(a | q' \right)$.

\subsection{Training and Inference}

We use a log-likelihood objective for training, which maximizes the
likelihood of the correct answer given a question:
\begin{equation}
\label{eq:objective}
\maximize \sum_{(q , a) \in \mathcal{D} }{ \log{p \left( a | q \right)}}
\end{equation}
where $\mathcal{D}$ is the set of all question-answer training pairs,
and~$p \left( a | q \right)$ is computed as shown in
Equation~\eqref{eq:overview}.
For the knowledge base QA task, we predict how likely it is that a subgraph obtains the correct answer, and the answers of some candidate subgraphs are partially correct. So, we use the binary cross entropy between the candidate subgraph's F1 score and the prediction as the objective function.
The RMSProp algorithm~\cite{rmsprop} is
employed to solve this non-convex optimization problem. Moreover,
dropout is used for regularizing the recurrent neural
networks~\cite{rnn:dropout}.

At test time, we generate paraphrases for the question $q$, and then predict
the answer by:
\begin{equation}
\label{eq:inference}
\hat{a} = \argmax_{a' \in \mathcal{C}_q}{ p \left( a' | q \right) }
\end{equation}
where $\mathcal{C}_q$ is the set of candidate answers, and~$p \left(
  a' | q \right)$ is computed as shown in
Equation~\eqref{eq:overview}.

\section{Experiments}
We compared our model which we call \textsc{Para4QA} (as shorthand for
learning to paraphrase for question answering) against multiple
previous systems on three datasets. In the following we introduce
these datasets, provide implementation details for our model, describe
the systems used for comparison, and present our results.

\subsection{Datasets}
Our model was trained on three datasets, representative of different
types of QA tasks.  The first two datasets focus on question answering
over a structured knowledge base, whereas the third one is specific to
answer sentence selection.

\paragraph{\textsc{WebQuestions}} This dataset~\cite{webqa} contains
$3,778$~training instances and $2,032$~test instances. Questions were
collected by querying the Google Suggest API. A breadth-first search
beginning with \textit{wh-} was conducted and the answers were
crowd-sourced using Freebase as the backend knowledge
base.

\paragraph{\textsc{GraphQuestions}} The dataset~\cite{graphqa}
contains~$5,166$ question-answer pairs (evenly split into a training
and a test set). It was created by asking crowd workers to paraphrase
$500$~Freebase graph queries in natural language.

\paragraph{\textsc{WikiQA}}
This dataset~\cite{wikiqa} has $3,047$~questions sampled from Bing
query logs. The questions are associated with $29,258$~candidate
answer sentences, $1,473$~of which contain the correct answers to the
questions.

\subsection{Implementation Details}

\paragraph{Paraphrase Generation}
Candidate paraphrases were stemmed~\cite{stemmer} and lowercased.
We discarded duplicate or trivial paraphrases which only rewrite stop words or punctuation.
For the NMT model, we followed the
implementation\footnote{\url{github.com/sebastien-j/LV_groundhog}} and
settings described in~\citet{jnmt}, and used
English$\leftrightarrow$German as the language pair.  The system was
trained on data released as part of the WMT15 shared translation task
($4.2$~million sentence pairs). We also had access to back-translated
monolingual training data~\cite{nmt:monolingual:data}.  Rare words
were split into subword units~\cite{nmt:subword} to handle
out-of-vocabulary words in questions. We used the top~$15$ decoding
results as candidate paraphrases.
We used the S size package of PPDB 2.0~\cite{ppdb2} for high
precision. At most $10$~candidate paraphrases were considered.
We mined paraphrase rules from WikiAnswers \cite{openqa14} as
described in Section~\ref{sec:para:rule}. The extracted rules were
ranked using the pointwise mutual information between template pairs
in the {WikiAnswers} corpus.  The top $10$ candidate paraphrases were
used.

\paragraph{Training}
For the paraphrase scoring model, we used GloVe~\cite{glove}
vectors\footnote{\url{nlp.stanford.edu/projects/glove}} pretrained on
Wikipedia 2014 and Gigaword 5 to initialize the word embedding
matrix. We kept this matrix fixed across datasets.
Out-of-vocabulary words were replaced with a special unknown
symbol. We also augmented questions with start-of- and end-of-sequence
symbols. Word vectors for these special symbols were updated during
training.  Model hyperparameters were validated on the development
set. The dimensions of hidden vectors and word embeddings were
selected from $\{50,100,200\}$ and $\{100,200\}$, respectively.
The dropout rate was selected from $\{0.2,0.3,0.4\}$.
The \textsc{BiLSTM} for the answer sentence selection QA model used
the same hyperparameters.
Parameters were randomly initialized from a
uniform distribution $\mathcal{U}\left(-0.08,0.08\right)$.
The learning rate and decay rate of RMSProp were $0.01$ and $0.95$, respectively.
The batch size was set to $150$. To alleviate the
exploding gradient problem~\cite{pascanu2013difficulty}, the gradient
norm was clipped to~$5$. Early stopping was used to determine the
number of epochs.

\subsection{Paraphrase Statistics}
\label{sec:paraphr-stat}

\begin{table}[t]
	\centering
	\setlength\tabcolsep{1.5pt}
	\fontsize{7}{8.5}\selectfont
	\begin{tabular}{lllllllllllll}
		\toprule
		\textbf{Metric}           &
                                            \multicolumn{3}{c}{\textsc{\textbf{GraphQ}}} & &  \multicolumn{3}{c}{\textsc{\textbf{WebQ
}}} & &  \multicolumn{3}{c}{\textsc{\textbf{WikiQA}}} \\ \cmidrule{2-4} \cmidrule{6-8} \cmidrule{10-12} 
		& \textbf{NMT} & \textbf{PPDB} & \textbf{Rule} & & \textbf{NMT} & \textbf{PPDB} & \textbf{Rule} & &\textbf{NMT} & \textbf{PPDB} & \textbf{Rule} \\ \hline
		$\avg(|q|)$ & & 10.87 & & & & 7.71 & & & & 6.47 & \\
		$\avg(|q'|)$ & 10.87 & 12.40 & 10.51 & & 8.13 & 8.55 & 7.54 & & 6.60 & 7.85 & 7.15 \\
		$\avg(\#q')$ & 13.85 & 3.02 & 2.50 & & 13.76 & 0.71 & 7.74 & & 13.95 & 0.62 & 5.64 \\
		Coverage (\%) & 99.67 & 73.52 & 31.16 & & 99.87 & 35.15 & 83.61  & & 99.89 & 31.04 & 63.12 \\
		BLEU (\%) & 42.33 & 67.92 & 54.23 & & 35.14 & 56.62 & 42.37 & &  32.40 & 54.24 & 40.62 \\
		TER (\%) & 39.18 & 14.87 & 38.59 & & 45.38 & 19.94 & 43.44 & &  46.10 & 17.20 & 48.59 \\
		\bottomrule
	\end{tabular}
	\caption{Statistics of generated paraphrases across datasets
		(training set). $\avg(|q|)$: average question
		length; $\avg(|q'|)$: average paraphrase length; $\avg(\#q')$:
		average number of paraphrases; coverage: the proportion of
		questions that have at least one candidate paraphrase.} 
	\label{table:results:statistics}
\end{table}

Table~\ref{table:results:statistics} presents descriptive statistics
on the paraphrases generated by the various systems across datasets
(training set).  As can be seen, the average paraphrase length is
similar to the average length of the original questions.  The NMT
method generates more paraphrases and has wider coverage, while the
average number and coverage of the other two methods varies per
dataset.  As a way of quantifying the extent to which rewriting takes
place, we report BLEU~\cite{bleu} and TER~\cite{ter} scores between
the original questions and their paraphrases.  The NMT method and the
rules extracted from WikiAnswers tend to paraphrase more (i.e., have
lower BLEU and higher TER scores) compared to PPDB.

\begin{table}[t]
	\small
	\centering
	\setlength\tabcolsep{1.5pt}
	\begin{tabular}{p{4.75cm} l l}
		\toprule
		\textbf{Method}                               &    \multicolumn{2}{c}{\textbf{Average F1} (\%)}     \\
		\cmidrule{2-3}                                & \textbf{\textsc{GraphQ}} & \textbf{\textsc{WebQ}} \\ \hline
		\textsc{Sempre}~\cite{webqa}                  & 10.8                      & 35.7                    \\
		\textsc{Jacana}~\cite{jacana}                 & 5.1                       & 33.0                    \\
		\textsc{ParaSemp}~\cite{parasempre}           & 12.8                      & 39.9                    \\
		\textsc{SubGraph}~\cite{qa:subgraph}          & -                         & 40.4                    \\
		\textsc{MCCNN}~\cite{qa:mccnn}                & -                         & 40.8                    \\
		\textsc{Yao15}~\cite{yao:2015:demos}          & -                         & 44.3                    \\
		\textsc{AgendaIL}~\cite{AgendaIL}             & -                         & 49.7                    \\
		\textsc{STAGG}~\cite{stagg}                   & -                         & 48.4 (52.5)             \\
		\textsc{MCNN}~\cite{mcnn}                     & -                         & 47.0 (53.3)             \\
		\textsc{TypeRerank}~\cite{sp:type:inference}  & -                         & \textbf{51.6} (52.6)    \\
		\textsc{BiLayered}~\cite{latent-pcfg-sp}      & -                         & 47.2                    \\
		\textsc{UDepLambda}~\cite{reddy2017universal} & \textbf{17.6}             & 49.5                    \\ \hline
		\textsc{SimpleGraph} (baseline)               & 15.9                      & 48.5                    \\
		\textsc{AvgPara}                              & 16.1                      & 48.8                    \\
		\textsc{SepPara}                              & 18.4                      & 49.6                    \\
		\textsc{DataAugment}                          & 16.3                      & 48.7                    \\
		\textsc{Para4QA}                              & \textbf{20.4}             & \textbf{50.7}           \\
		~~$-$\textsc{NMT}                             & 18.5                      & 49.5                    \\
		~~$-$\textsc{PPDB}                            & 19.5                      & 50.4                    \\
		~~$-$\textsc{Rule}                            & 19.4                      & 49.1                    \\ \bottomrule
	\end{tabular}
	\caption{Model performance on \textsc{GraphQuestions} and
		\textsc{WebQuestions}. Results with additional task-specific
		resources are shown in parentheses. The base QA model is
		\textsc{SimpleGraph}. Best results in each group are shown in \textbf{bold}.}
	\label{table:results:graphqa:webqa}
\end{table}

\subsection{Comparison Systems}
\label{sec:comparison-systems}

We compared our framework to previous work and several ablation models
which either do not use paraphrases or paraphrase scoring, or are not
jointly trained. 

The first baseline only uses the base QA models described in
Section~\ref{sec:qa:model} (\textsc{SimpleGraph} and
\textsc{BiLSTM}). The second baseline (\textsc{AvgPara}) does not take
advantage of paraphrase scoring. The paraphrases for a given question
are used while the QA model's results are directly averaged to predict
the answers. The third baseline (\textsc{DataAugment}) employs
paraphrases for data augmentation during training. Specifically, we
use the question, its paraphrases, and the correct answer to
automatically generate new training samples.

In the fourth baseline (\textsc{SepPara}), the paraphrase scoring
model is separately trained on paraphrase classification data, without
taking question-answer pairs into account.  In the experiments, we
used the Quora question paraphrase
dataset\footnote{\url{goo.gl/kMP46n}} which contains question pairs
and labels indicating whether they constitute paraphrases or not. We
removed questions with more than $25$~tokens and sub-sampled to
balance the dataset.  We used $90\%$ of the resulting $275$K~examples
for training, and the remaining for development.  The paraphrase score
$s\left(q' | q\right)$ (Equation~\eqref{eq:score}) was wrapped by a
$\sigmoid$ function to predict the probability of a question pair
being a paraphrase. A binary cross-entropy loss was used as the
objective. The classification accuracy on the dev set was~$80.6\%$.

Finally, in order to assess the individual contribution of different
paraphrasing resources, we compared the \textsc{Para4QA} model against
versions of itself with one paraphrase generator removed
($-$\textsc{NMT}/$-$\textsc{PPDB}/$-$\textsc{Rule}).

\subsection{Results}

\begin{table}[t]
	\small
	\centering
	\setlength\tabcolsep{3pt}
	\begin{tabular}{l l l}
		\toprule
		\textbf{Method}                         & \textbf{MAP}    & \textbf{MRR}    \\ \hline
		\textsc{BigramCNN}~\cite{bigramcnn}     & 0.6190          & 0.6281          \\
		\textsc{BigramCNN+cnt}~\cite{bigramcnn} & 0.6520          & 0.6652          \\
		\textsc{ParaVec}~\cite{pv}              & 0.5110          & 0.5160          \\
		\textsc{ParaVec+cnt}~\cite{pv}          & 0.5976          & 0.6058          \\
		\textsc{LSTM}~\cite{nasm}               & 0.6552          & 0.6747          \\
		\textsc{LSTM+cnt}~\cite{nasm}           & 0.6820          & 0.6988          \\
		\textsc{NASM}~\cite{nasm}               & 0.6705          & 0.6914          \\
		\textsc{NASM+cnt}~\cite{nasm}           & 0.6886          & 0.7069          \\
		\textsc{KvMemNet+cnt}~\cite{kv:memnet}  & \textbf{0.7069} & \textbf{0.7265} \\ \hline
		\textsc{BiLSTM} (baseline)              & 0.6456          & 0.6608          \\
		\textsc{AvgPara}                        & 0.6587          & 0.6753          \\
		\textsc{SepPara}                        & 0.6613          & 0.6765          \\
		\textsc{DataAugment}                    & 0.6578          & 0.6736          \\
		\textsc{Para4QA}                        & \textbf{0.6759} & \textbf{0.6918} \\
		~~$-$\textsc{NMT}                       & 0.6528          & 0.6680          \\
		~~$-$\textsc{PPDB}                      & 0.6613          & 0.6767          \\
		~~$-$\textsc{Rule}                      & 0.6553          & 0.6756          \\ \hline
		\textsc{BiLSTM+cnt} (baseline)          & 0.6722          & 0.6877          \\
		\textsc{Para4QA+cnt}                    & \textbf{0.6978} & \textbf{0.7131} \\ \bottomrule
	\end{tabular}
	\caption{Model performance on \textsc{WikiQA}. \textsc{+cnt}: 
		word matching features introduced in~\citet{wikiqa}. The base QA model is \textsc{BiLSTM}. Best results in each
		group are shown in \textbf{bold}.} 
	\label{table:results:wikiqa}
\end{table}

We first discuss the performance of \textsc{Para4QA} on
\textsc{GraphQuestions} and \textsc{WebQuestions}.  The first block in
Table~\ref{table:results:graphqa:webqa} shows a
variety of systems previously described in the literature using
average F1 as the evaluation metric~\cite{webqa}. Among these,
\textsc{ParaSemp}, \textsc{SubGraph}, \textsc{MCCNN}, and
\textsc{BiLayered} utilize paraphrasing resources.
The second block compares \textsc{Para4QA} against various related
baselines (see Section~\ref{sec:comparison-systems}).
\textsc{SimpleGraph} results on \textsc{WebQuestions} and
\textsc{GraphQuestions} are taken from~\citet{depgraphparser} and
\citet{reddy2017universal}, respectively.

Overall, we observe that \textsc{Para4QA} outperforms baselines which
either do not employ paraphrases (\textsc{SimpleGraph}) or paraphrase
scoring (\textsc{AvgPara}, \textsc{DataAugment}), or are not jointly
trained (\textsc{SepPara}). On \textsc{GraphQuestions}, our model
\textsc{Para4QA} outperforms the previous state of the art by a wide
margin.  Ablation experiments with one of the paraphrase generators
removed show that performance drops most when the NMT paraphrases are
not used on \textsc{GraphQuestions}, whereas on \textsc{WebQuestions}
removal of the rule-based generator hurts performance most.  One
reason is that the rule-based method has higher coverage on
\textsc{WebQuestions} than on \textsc{GraphQuestions} (see
Table~\ref{table:results:statistics}).

Results on \textsc{WikiQA} are shown in
Table~\ref{table:results:wikiqa}. We report MAP and MMR which evaluate
the relative ranks of correct answers among the candidate sentences
for a question.  Again, we observe that \textsc{Para4QA} outperforms
related baselines (see \textsc{BiLSTM}, \textsc{DataAugment},
\textsc{AvgPara}, and \textsc{SepPara}).  Ablation experiments show
that performance drops most when NMT paraphrases are removed.  When
word matching features are used (see \textsc{+cnt} in the third
block), \textsc{Para4QA} reaches state of the art performance. 


\begin{table}[t]
\fontsize{8.5}{9.5}\selectfont
\centering
\begin{tabular}{l l}
	\toprule
	\textbf{Examples}                                                             & ${p_{\theta}\left(q' | q\right)}$ \\ \hline
	\multicolumn{2}{l}{(\texttt{music.concert\_performance.performance\_role})}                                       \\
	\underline{\textcolor{myred}{\it what sort of part do queen play in concert}} & 0.0659                            \\
	what role do queen play in concert                                            & 0.0847                            \\
	what be the role play by the queen in concert                                 & 0.0687                            \\
	what role do queen play during concert                                        & 0.0670                            \\
	\textcolor{myred}{\it what part do queen play in concert}                     & 0.0664                            \\
	which role do queen play in concert concert                                   & 0.0652                            \\ \hline
	\multicolumn{2}{l}{(\texttt{sports.sports\_team\_roster.team})}                                                   \\
	\underline{\textcolor{myred}{\it what team do shaq play 4}}                   & 0.2687                            \\
	what team do shaq play for                                                    & 0.2783                            \\
	which team do shaq play with                                                  & 0.0671                            \\
	which team do shaq play out                                                   & 0.0655                            \\
	\textcolor{myred}{\it which team have you play shaq}                          & 0.0650                            \\
	what team have we play shaq                                                   & 0.0497                            \\ \bottomrule
\end{tabular}
\caption{Questions and their top-five
paraphrases with probabilities learned by the model.
The Freebase relations used to query the correct answers are shown in brackets.
The original question is \underline{underlined}.
Questions with incorrect predictions are in \textcolor{myred}{\it red}.}
\label{table:experiment:example}
\end{table}

Examples of paraphrases and their
probabilities~${p_{\theta}\left(q' | q\right)}$ (see
Equation~\eqref{eq:normalization}) learned by \textsc{Para4QA} are
shown in Table~\ref{table:experiment:example}. The two examples are
taken from the development set of \textsc{GraphQuestions} and
\textsc{WebQuestions}, respectively. We also show the Freebase
relations used to query the correct answers.  In the first example,
the original question cannot yield the correct answer because of the
mismatch between the question and the knowledge base.  The paraphrase
contains \tx{role} in place of \tx{sort of part}, increasing the
chance of overlap between the question and the predicate words.  The
second question contains an informal expression \tx{play 4}, which
confuses the QA model. The paraphrase model generates \tx{play for}
and predicts a high paraphrase score for it.  More generally, we
observe that the model tends to give higher
probabilities~${p_{\theta}\left(q' | q\right)}$ to paraphrases biased
towards delivering appropriate answers.


We also analyzed which structures were mostly paraphrased within a
question. We manually inspected $50$~(randomly sampled) questions from
the development portion of each dataset, and their three top scoring
paraphrases (Equation~\eqref{eq:score}).  We grouped the most commonly
paraphrased structures into the following categories: a) question
words, i.e., wh-words and and \tx{how}; b) question focus structures,
i.e., cue words or cue phrases for an answer with a specific entity
type~\cite{jacana}; c) verbs or noun phrases indicating the relation
between the question topic entity and the answer; and d)~structures
requiring aggregation or imposing additional constraints the answer
must satisfy~\cite{stagg}.  In the example \tx{which year did Avatar
  release in UK}, the question word is \tx{which}, the question focus
is \tx{year}, the verb is \tx{release}, and \tx{in UK} constrains the
answer to a specific location.

\begin{figure}[t]
	\centering
	\includegraphics[width=0.44\textwidth]{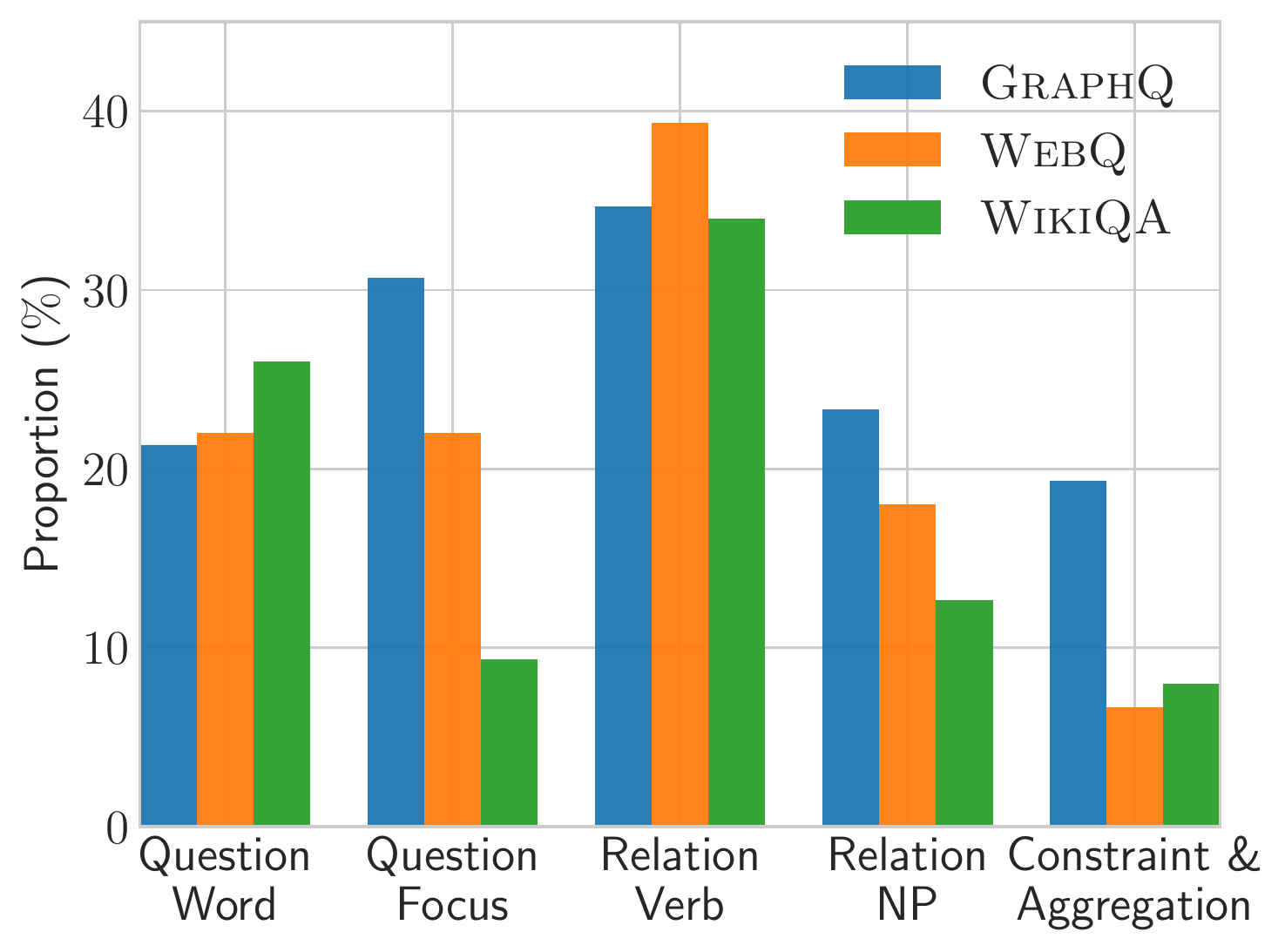}
	\vspace*{-2.5ex}
	\caption{Proportion of linguistic phenomena subject to
		paraphrasing within a question.}
	\label{fig:paraphrase:type}
\end{figure}

Figure~\ref{fig:paraphrase:type} shows the degree to which different
types of structures are paraphrased.  As can be seen, most rewrites
affect Relation Verb, especially on \textsc{WebQuestions}.  Question
Focus, Relation NP, and Constraint \& Aggregation are more often
rewritten in \textsc{GraphQuestions} compared to the other datasets.

\begin{table}[t]
	\small
	\centering
	\begin{tabular}{l l l}
		\toprule
		\textbf{Method}      &    \multicolumn{2}{c}{\textbf{Average F1} (\%)}     \\
		\cmidrule{2-3}       & \textbf{Simple} & \textbf{\tabincell{l}{Complex}} \\ \hline
		\textsc{SimpleGraph} & 20.9                      & 12.2                    \\
		\textsc{Para4QA}     & \textbf{27.4} (+6.5)      & \textbf{16.0} (+3.8)    \\ \bottomrule
	\end{tabular}
	\caption{We group \textsc{GraphQuestions} into simple
          and complex questions and report model performance in each
          split. Best results in each group are shown in \textbf{bold}. The values in brackets are absolute improvements of average F1 scores.}
	\label{table:results:simple_complex}
\end{table}

Finally, we examined how our method fares on simple versus complex
questions. We performed this analysis on \textsc{GraphQuestions} as it
contains a larger proportion of complex questions. We consider
questions that contain a single relation as simple. Complex questions
have multiple relations or require aggregation.
Table~\ref{table:results:simple_complex} shows how our model performs
in each group. We observe improvements for both types of questions,
with the impact on simple questions being more pronounced. This is not
entirely surprising as it is easier to generate paraphrases and predict
the paraphrase scores for simpler questions.

\section{Conclusions}
\label{sec:conclusions}

In this work we proposed a general framework for learning paraphrases
for question answering. Paraphrase scoring and QA models are trained
end-to-end on question-answer pairs, which results in learning
paraphrases with a purpose.  The framework is not tied to a specific
paraphrase generator or QA system. In fact it allows to incorporate
several paraphrasing modules, and can serve as a testbed for exploring
their coverage and rewriting capabilities.  Experimental results on
three datasets show that our method improves performance across tasks.
There are several directions for future work.  The framework can be
used for other natural language processing tasks which are sensitive
to the variation of input (e.g.,~textual entailment or summarization).
We would also like to explore more advanced paraphrase scoring models~\cite{decomposable:attention,mlstm} as well as additional paraphrase generators since
improvements in the diversity and the quality of paraphrases could
also enhance QA performance.

\paragraph{Acknowledgments}
The authors gratefully acknowledge the support of the UK Engineering
and Physical Sciences Research Council (grant EP/L016427/1; Mallinson)
and the European Research Council (award number 681760; Lapata).

\bibliography{para4qa}
\bibliographystyle{emnlp_natbib}

\end{document}